\documentclass[letterpaper]{article} 
\usepackage{aaai24}  
\usepackage{times}  
\usepackage{helvet}  
\usepackage{courier}  
\usepackage[hyphens]{url}  
\usepackage{graphicx} 
\usepackage{natbib}  
\usepackage{caption} 
\usepackage{algorithm}
\usepackage{algorithmic}

\usepackage{newfloat}
\usepackage{listings}
\DeclareCaptionStyle{ruled}{labelfont=normalfont,labelsep=colon,strut=off} 
\floatstyle{ruled}
\newfloat{listing}{tb}{lst}{}
\floatname{listing}{Listing}
\usepackage{array}
\usepackage{subfig}
\usepackage{multirow}
\usepackage{pgfplots}
\usepackage{graphicx}
\usepackage{paralist}
\usepackage{colortbl}
\usepackage{amsmath}
\usepackage{amssymb}
\pgfplotsset{compat=1.18}
\usepackage{afterpage}
\newcommand{\ignorethis}[1]{}

\newcommand{\orange}[1]{\textcolor[RGB]{234, 112,  13}{#1}} 

\newcommand{\dw}[1]{\textcolor[RGB]{63, 163, 77}{#1}} 

\title{Exploiting Polarized Material Cues for Robust Car Detection}

\author {
    Wen Dong\textsuperscript{\rm 1},
    Haiyang Mei\textsuperscript{\rm 1,2},
    Ziqi Wei\textsuperscript{\rm 3,4,$\ast$},
    Ao Jin\textsuperscript{\rm 1},
    Sen Qiu\textsuperscript{\rm 1},
    Qiang Zhang\textsuperscript{\rm 1},
    Xin Yang\textsuperscript{\rm 1,}\thanks{Corresponding authors}
}
\affiliations {
    \textsuperscript{\rm 1}Key Laboratory of Social Computing and Cognitive Intelligence, Dalian University of Technology\\
    \textsuperscript{\rm 2}Show Lab, National University of Singapore\\
    \textsuperscript{\rm 3}Institute of Automation, Chinese Academy of Sciences\\
    \textsuperscript{\rm 4}State Key Laboratory of Structural Analysis for Industrial Equipment, Dalian University of Technology\\
    \{dongwen, dllsja\}@mail.dlut.edu.cn, haiyang.mei@outlook.com, ziqi.wei@ia.ac.cn, \{qiu, zhangq, xinyang\}@dlut.edu.cn
}

\begin{document}
\maketitle

\begin{figure*}[ht]
  \centering
  \includegraphics[width=\textwidth]{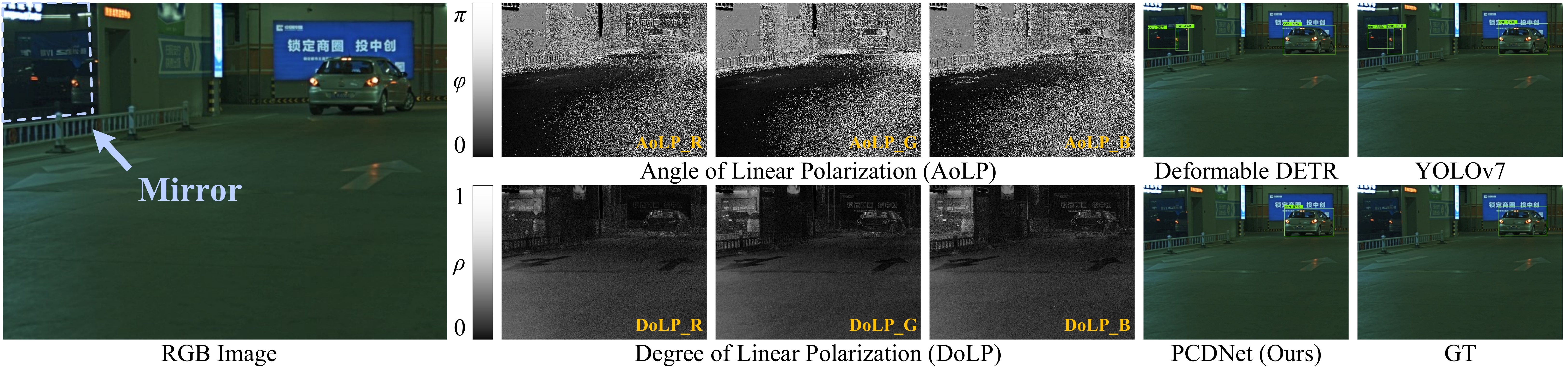}
  \caption{Car detections (indicated by \dw{green} bounding boxes) obtained with the \textit{RGB-only} methods of Deformable DETR \cite{zhu2020deformable} and YOLOv7 \cite{wang2022yolov7} compared to our \textit{RGB-Polarization} Car Detection Network (PCDNet). Both prior methods fail to distinguish mirrored cars from real ones due to the similar visual appearances. In contrast, our method can handle such ambiguity and correctly detect the real car in scenes with the help of intrinsic material properties revealed by the polarization cues.}
  \label{fig:teaser}
\end{figure*}

\begin{abstract}
Car detection is an important task that serves as a crucial prerequisite for many automated driving functions. The large variations in lighting/weather conditions and vehicle densities of the scenes pose significant challenges to existing car detection algorithms to meet the highly accurate perception demand for safety, due to the unstable/limited color information, which impedes the extraction of meaningful/discriminative features of cars. In this work, we present a novel learning-based car detection method that leverages trichromatic linear polarization as an additional cue to disambiguate such challenging cases. A key observation is that polarization, characteristic of the light wave, can robustly describe intrinsic physical properties of the scene objects in various imaging conditions and is strongly linked to the nature of materials for cars (\textit{e.g.}, metal and glass) and their surrounding environment (\textit{e.g.}, soil and trees), thereby providing \textbf{\textit{reliable}} and \textbf{\textit{discriminative}} features for robust car detection in challenging scenes. To exploit polarization cues, we first construct a pixel-aligned RGB-Polarization car detection dataset, which we subsequently employ to train a novel multimodal fusion network. Our car detection network dynamically integrates RGB and polarization features in a request-and-complement manner and can explore the intrinsic material properties of cars across all learning samples. We extensively validate our method and demonstrate that it outperforms state-of-the-art detection methods. Experimental results show that polarization is a powerful cue for car detection. Our code is available at \url{https://github.com/wind1117/AAAI24-PCDNet}.
\end{abstract}

\section{Introduction}
\label{sec:introduction}
Autonomous driving and advanced driving assistance system (ADAS) rely on highly accurate road scene analysis. As a fundamental step to achieving a reliable road scene understanding, object detection has received great attention in recent years and has been significantly boosted with the development of deep neural networks (DNNs) \cite{lin2017focal, redmon2018yolov3, sun2021sparse}. In practical road scenes, cars appear to be one of the most frequently observed yet dangerous objects, and car detection is still a challenging problem due to the large structural and appearance variations of cars in different scenes.

Although existing state-of-the-art methods explore rich contexts in multiple modalities including RGB, LiDAR, and infrared to improve the detection accuracy, these methods typically assume the imaging quality is optimal. When it comes to adverse conditions such as low light, rain, and fog, a significant accuracy drop would occur due to the poor and limited sensing scene information fed into the algorithms. For example, an RGB camera may fail to capture important visual cues under low-light conditions \cite{song2019vision, arora2022automatic}, a LiDAR sensor may struggle to distinguish targets in complex environments due to its limited range and resolution \cite{Qian_2021_CVPR, chen2020pseudo}, and, similarly, an infrared sensor may produce blurry images with low contrast when exposed to extreme weather conditions \cite{du2021weak, sun2022drone}. Instead, polarization, characteristic of the light wave, can robustly reveal the intrinsic physical properties of cars and their surrounding environment (\textit{e.g.}, the surface geometric structure, roughness, and material) in various view/lighting/weather conditions. This inspires us to exploit the \textbf{\textit{reliable}} and \textbf{\textit{discriminative}} features provided by polarization to complement traditional RGB features for robust car detection.

Linear polarization cues, described by the angle of polarization (AoLP) and the degree of linear polarization (DoLP), might not be equally obvious/informative over different scenes and image regions, or even confound valid RGB cues. To address these challenges, we design a novel RGB-Polarization Car Detection Network (PCDNet) with RGB intensities, trichromatic AoLP and DoLP as input.
PCDNet is built on three key modules: (i) Polarization Integration (PI) module that fuses AoLP and DoLP to generate a comprehensive and semantically meaningful polarization representation; (ii) Material Perception (MP) module to explore the polarization/material properties of cars across different learning samples for enhancing the polarization cues in each scene; and (iii) Cross-Domain Demand Query (CDDQ) module to dynamically integrate the informative polarization cues into RGB features based on the spatial demand map generated from RGB domain.

To train PCDNet, we introduce an RGB-Polarization car detection dataset, dubbed RGBP-Car, which consists of 1,611 RGB images and pixel-aligned trichromatic (\textit{i.e.}, red, green and blue channels) AoLP and DoLP images, as well as corresponding annotated 31,234 bounding boxes of cars. To ensure diversity, the images in RGB-P Car are captured from various real-world traffic scenes with different view/weather/lighting conditions and vehicle densities.

We perform extensive experiments to demonstrate the superiority of our method over competing approaches and show the importance of polarization cues for robust car detection in challenging scenes (\textit{e.g.}, Fig. \ref{fig:teaser}). In summary, our contributions are:
\begin{itemize}
\item the first solution to exploit both RGB and trichromatic angle/degree of linear polarization (AoLP/DoLP) cues for robust car detection;
\item a new pixel-aligned RGB-P car detection dataset covering challenging dense cars and low light scenarios;
\item a novel multimodal fusion network that dynamically integrates RGB and polarization features in a request-and-complement manner;
\item a novel polarization cues perception strategy to implicitly explore the intrinsic material properties of cars across the whole learning samples.
\end{itemize}

\section{Background and Related Work}
\label{sec:related_work}

\textbf{Polarization} has a long research history in computer vision and is widely used in many tasks such as reflection removal \cite{wieschollek2018separating, lei2020polarized, li2020reflection}, surface normal and/or shape estimation \cite{chen2017multi, kadambi2015polarized}, and semantic segmentation \cite{mei2022glass, kalra2020deep}. Light is an electromagnetic wave, with its electric field oscillating perpendicularly to the direction of propagation. Unpolarized light has a randomly fluctuating electric field while polarized light has a biased direction of the electric field. Common light sources like the sun and LED spotlights emit unpolarized light which would become partially/fully polarized light when passing through a linear polarizer, reflecting off certain materials, or undergoing certain types of scattering.

In this work, we focus on the linear polarization measurement captured by the off-the-shelf polarization-array CMOS sensor which can record light intensities in four polarization directions, \textit{i.e.}, $I_{0^{\circ}}$, $I_{45^{\circ}}$, $I_{90^{\circ}}$, and $I_{135^{\circ}}$, respectively. The polarization state of the light can be described using the Stokes vector $S=[S_0, S_1, S_2, S_3]$, where $S_0$ stands for the total light intensity, $S_1$ and $S_2$ describe the ratio of the $0^{\circ}/45^{\circ}$ linear polarization over its perpendicular counterpart, and $S_3$ is the circular polarization power. The Stokes elements $S_0, S_1, S_2$ are formally defined as:
\begin{align} 
\label{eq:stokes}
\begin{array}{lr}
    S_0 =I_{0^{\circ}} + I_{90^{\circ}} = I_{45^{\circ}} + I_{135^{\circ}},\\
    S_1 =I_{0^{\circ}}-I_{90^{\circ}},\\
    S_2 =I_{45^{\circ}}-I_{135^{\circ}}.
\end{array}
\end{align}

The angle of linear polarization (AoLP) $\phi$ and the degree of linear polarization (DoLP) $\rho$ are then be calculated via:
\begin{align} 
\label{eq:polar}
    \phi = \frac{1}{2}arctan(\frac{S_2}{S_1}), \quad \rho = \frac{\sqrt{S_1^2+S_2^2}}{S_0}.
\end{align}

As shown in Fig. \ref{fig:samples}, the trees, walls or sky in the background usually exhibit a low linear polarization degree while the glass, rubber and plastic parts of a car are typically with a high linear polarization degree. This observation inspires us to exploit the material cues revealed by polarization for robust car detection.

\begin{figure*}[ht]
    \centering
    \includegraphics[width=\linewidth]{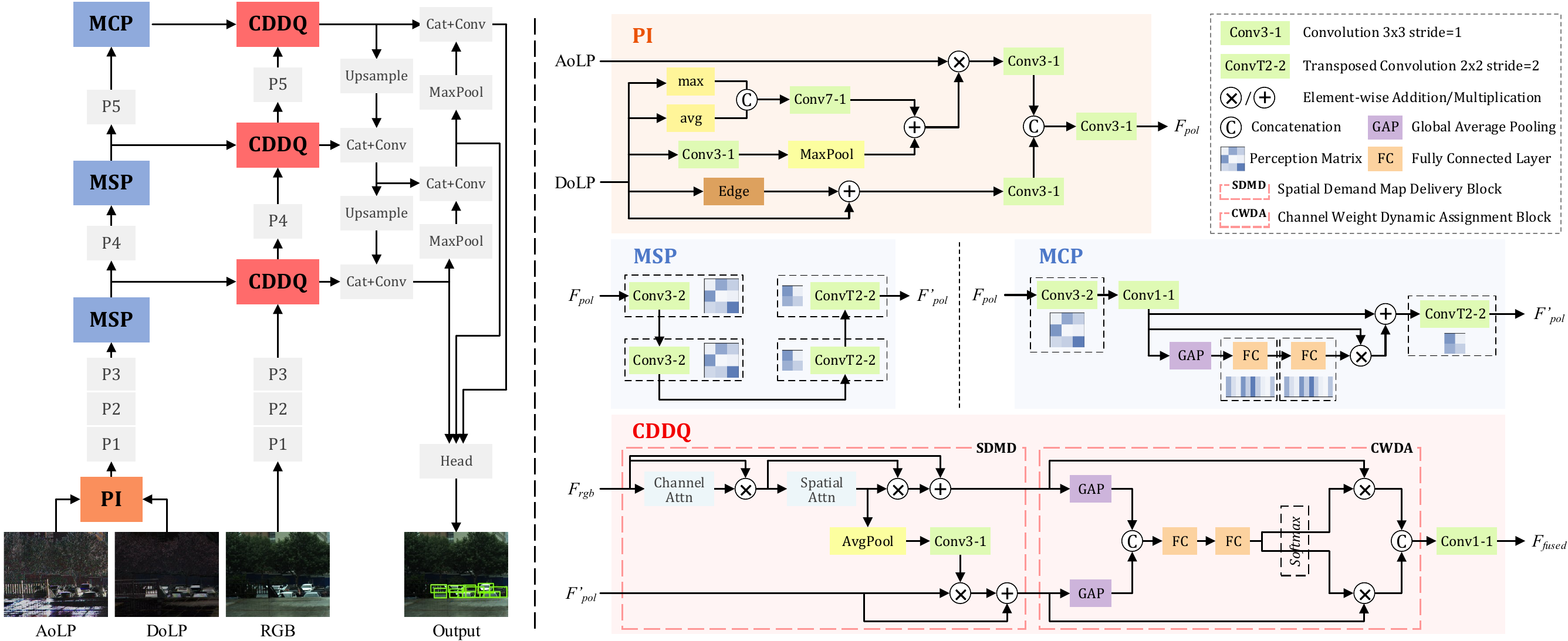}
    \caption{Overview of PCDNet and its three main modules: the Polarization Integration (PI) module, the Material Spatial/Channel Perception (MSP/MCP) module, and the Cross Domain Demand Query (CDDQ) module.}
    \label{fig:pipeline}
\end{figure*}

\textbf{Object Detection} has achieved significant progress with the revolution of deep learning. Many state-of-the-art approaches emerged, including region-based detectors (\textit{e.g.}, Faster R-CNN \cite{Ren_2017} and EfficientDet \cite{tan2020efficientdet}), one-stage detectors (\textit{e.g.}, YOLO \cite{redmon2016you}, SSD \cite{liu2016ssd}, and RetinaNet \cite{lin2017focal}), and anchor-free detectors (\textit{e.g.}, FCOS \cite{tian2019fcos}). These methods typically employ advanced network architectures such as ResNet \cite{he2016deep}, VGG \cite{simonyan2014very} and EfficientNet \cite{tan2019efficientnet}. The trend in object detection has been toward the development of large models. The Vision Transformer \cite{dosovitskiy2020image}, derived from natural language processing \cite{zaheer2020big}, has achieved notable improvements in the field of object detection. For example, some methods such as Swin Transformer \cite{liu2021swin}, DETR \cite{carion2020end}, and DAB-DETR \cite{liu2022dabdetr} have achieved remarkable results on benchmark datasets including Microsoft COCO \cite{lin2014microsoft} and PASCAL VOC \cite{everingham2010pascal}. However, such models often exceed the hardware load and detection speed requirements of most restricted terminal devices. Moreover, most of them rely on clear and optimal RGB images which is hard to obtain in challenging scenes. Image enhancement and restoration on the low-quality RGB image will cost extra computing power and time. Our method differs from the above works in that we introduce reliable polarized material cues to complement traditional RGB features and design an RGB-P-based multimodal fusion network for robust detection.

\textbf{Multimodal Fusion} can provide rich contextual information for robust object detection \cite{valverde2021there, bijelic2020seeing}. Blin et al. employed a simple fusion method by stacking the multimodal data in the channel dimension to replace the original input \cite{blin2019road}. Manjunath et al. and Chen et al. adopted the concatenation \cite{manjunath2018radar} and element-wise addition \cite{chen2017multi} to fusing the low-level features of LiDAR and RGB, respectively. The attention mechanism \cite{vaswani2017attention} is also used to achieve multimodal fusion. HAFNet \cite{zhang2020hybrid} developed a cross-modal attention mechanism to perform feature fusion. Mei et al. calculated dynamic fusion weights for RGB and depth, considering the quality of each modality \cite{mei2021depth}. Similarly, Ji et al. used global average pooling followed by a fully connected layer to compute the channel attention weight for each modality \cite{ji2021calibrated}. Mei et al. generated spatial attention maps based on both global and local features to guide the multimodal fusion \cite{mei2022glass}. Although these methods achieve performance improvement to some extent, they perform information compensation in a passive ``post'' manner, resulting in the limited robustness of the model in challenging scenes. In this work, we develop a novel polarization material perception scheme to learn the intrinsic material properties of cars and a proactive multimodal fusion strategy to compensate RGB features with informative polarization cues in a ``request-and-complement'' manner, enhancing the robustness of car detection.

\section{Methodology}
\label{sec:methodology}
RGB images depict objects based on color differences that match human perception. However, objects with similar colors may not have enough color contrast to show their shape. By contrast, polarized light is strongly linked to the material of objects and the orientation of the reflecting surface, enabling it to reveal material properties that make object boundaries visible even when colors are similar (\textit{e.g.}, Fig. \ref{fig:samples} (a) and (b)). In spite of that, polarization cues may be weak in certain lighting conditions or viewing angles (\textit{e.g.}, Fig. \ref{fig:samples} (c)). Including polarization measurements naively in existing car detection methods may not necessarily yield the expected performance improvement. How to effectively integrate RGB and polarization features is a key challenge to be addressed to achieve robust car detection.

We introduce a novel Polarization Car Detection Network (PCDNet) that is capable of exploring and integrating polarized material cues for robust car detection. As shown in Fig. \ref{fig:pipeline}, PCDNet takes as input the RGB intensity, trichromatic AoLP and DoLP, and outputs car detections. The AoLP and DoLP are first integrated into a comprehensive and semantically meaningful polarization feature representation by a Polarization Integration (PI) module for the ease of the following feature extraction and fusion. Then, the RGB and polarization features are separately fed into two branches of CSPDarkNet \cite{wang2020cspnet} encoder, each consisting of five stages to extract multi-level contextual features. The Material Perception (MP) module, which aims to extract the intrinsic material properties of cars across the whole learning samples, is applied to different levels of extracted features. The MP module has a specific design for low-level features (MSP with spatial specialization) and high-level features (MCP with channel specialization), respectively. For multimodal feature fusion, the Cross-Domain Demand Query (CDDQ) module assigns fusion weights adaptively and conducts dynamic fusion in a request-and-complement manner. Finally, based on the fused features, we adopt the anchor-based detection head from YOLOv7 \cite{wang2022yolov7} to generate final classifications and bounding boxes.

\subsection{Polarization Integration (PI)}
AoLP $\phi$ and DoLP $\rho$ reveal object/scene materials from two different aspects. Polarization Integration (PI) module is designed to combine them into an unified and semantically meaningful polarization representation $F_{pol}$ for the ease of the following feature extraction and fusion. As the captured polarization angle is more likely random at regions with a low polarization degree, PI filters the AoLP measurement based on the DoLP. In addition, PI also extracts and integrates the edge information in DoLP measurement to help the distinction of objects with different materials. Formally,
\begin{align} \label{eq:pim}
    F_{pol} &=\vartheta_{3\cdot 1}([\vartheta_{3\cdot 1}(F_{\phi\rho}), \vartheta_{3\cdot 1}(\rho+\mathbb{E}(\rho))]), \\
    F_{\phi\rho} &=\phi\otimes(\vartheta_{3\cdot 1}([\hat{\mathbb{A}}(\rho),\hat{\mathbb{M}}(\rho)])+\sigma(\mathbb{M}(\vartheta_{3\cdot 1}(\rho)))),
\end{align}
where $\vartheta_{k\cdot s}$ denotes a $k \times k$ convolution with a stride of $s$, followed by a batch normalization and a SiLU activation function. $[\cdot]$ indicates the concatenation operation over the channel dimension. $\mathbb{E}$ refers to the Scharr edge extractor. $\otimes$ is the element-wise multiplication. $\hat{\mathbb{A}}$ and $\hat{\mathbb{M}}$ are the average and max pooling in the channel dimension, respectively. $\mathbb{M}$ is the max pooling in the spatial dimension with a kernel size of 5. And $\sigma$ is the Sigmoid activation.

\subsection{Material Perception (MP)}
Despite the cars in different scenarios may have diverse visual appearances such as different colors and textures, they are typically share similar materials including glass, rubber, metal, etc. Fortunately, polarization can robustly reveal the intrinsic physical properties of these materials. Inspired by this, we design the Material Perception (MP) strategy to explore
the discriminative and invariant material features of cars across different scenarios. Considering the different characteristics of different levels of features, \textit{i.e.}, low-level features have larger spatial sizes and keep rich and detailed low-level information while high-level features contain more semantic cues distributed in more feature channels, MP is instantialized as Material Spatial/Channel Perception (MSP/MCP) modules for the low-/high-level polarization features, respectively. Formally, MSP/MCP can be described as:
\begin{align} \label{eq:mpm}
    MSP(F) &=\varrho_{2\cdot2}(\varrho_{2\cdot2}(\vartheta_{3\cdot2}(\vartheta_{3\cdot2}(F)))), \\
    MCP(F) &=\varrho_{2\cdot2}(\mathcal{M}((\vartheta_{1\cdot1}(\vartheta_{3\cdot2}(F))))), \\
    \mathcal{M}(x) &=x+x\otimes\sigma(m_2(m_1(\bar{\mathbb{A}}(x)))),
\end{align}
where $\varrho_{k\cdot s}$ denotes a $k \times k$ transposed convolution with a stride of $s$, followed by a batch normalization and a SiLU activation function. $\mathcal{M}(\cdot)$ is the perception scheme with two independent perception matrices in fully connected layers named $m_1$ and $m_2$. And $\bar{\mathbb{A}}$ is global average pooling.

\subsection{Cross Domain Demand Query (CDDQ)}
Polarization and RGB features are different types of representations of the scenes and simply combing them may dilute the useful clues of the cars originally presented in the individual modality or amplify the background interference. We address this issue by introducing the Cross-Domain Demand Query (CDDQ) module for effective multimodal feature fusion, taking into account both the context and quality of each modality feature. CDDQ takes as input the RGB features $F_{rgb}$ and polarization representations $F'_{pol}$. It first utilizes a Spatial Demand Map Delivery (SDMD) block to generate enhanced RGB features $F_{rgb}^{*}$ and distill informative and required polarization features $F_{pol}^{*}$ and then obtains fused features $F_{fused}$ through a Channel Weight Dynamic Assignment (CWDA) block:
\allowdisplaybreaks\begin{align}
    F_{rgb}^{*} &=F_{rgb}+F''_{rgb} = F_{rgb}+\mu\otimes F'_{rgb} \nonumber \\
        &=F_{rgb}+\mu\otimes \eta\otimes F_{rgb}, \\
    \eta &=\sigma(\vartheta_{1\cdot1}(\bar{\mathbb{M}}(F_{rgb}))+\vartheta_{1\cdot1}(\bar{\mathbb{A}}(F_{rgb}))), \\
    \mu &=\sigma(\vartheta_{7\cdot1}([\hat{\mathbb{A}}(F'_{rgb}), \hat{\mathbb{M}}(F'_{rgb})])), \\
    F_{pol}^{*} &=F'_{pol}+\vartheta_{3\cdot1}(\mathbb{A}(\mu))\otimes F'_{pol}, \\
    F_{fused} &=\vartheta_{1\cdot1}([\alpha\times F_{rgb}^{*},\beta\times F_{pol}^{*}]), \\
    \alpha, \beta &=\delta(\langle\sigma(fc(si(fc([\bar{\mathbb{A}}(F_{rgb}^{*}), \bar{\mathbb{A}}(F_{pol}^{*})]))))\rangle),
\end{align}
where $\eta$ is the channel attention vector. $\mu$ is the spatial attention map which also serves as the guidance of the informative polarization cues request process. $\mathbb{A}$ is an average pooling with a kernel size of 3. $\bar{\mathbb{M}}$/$\bar{\mathbb{A}}$ denotes global max/average pooling. $\alpha$ and $\beta$ are dynamic fusion weights assigned to the RGB and polarization features, respectively, with a constraint of being non-negative and summing up to 1 for each channel position. $fc$ is the fully connected layer and $\langle\cdot\rangle$ is the split operation over the channel dimension. $si$ and $\delta$ are the SiLU and Softmax activation functions, respectively.

\begin{figure}[ht]
    \begin{center}
        \includegraphics[width=1\linewidth]{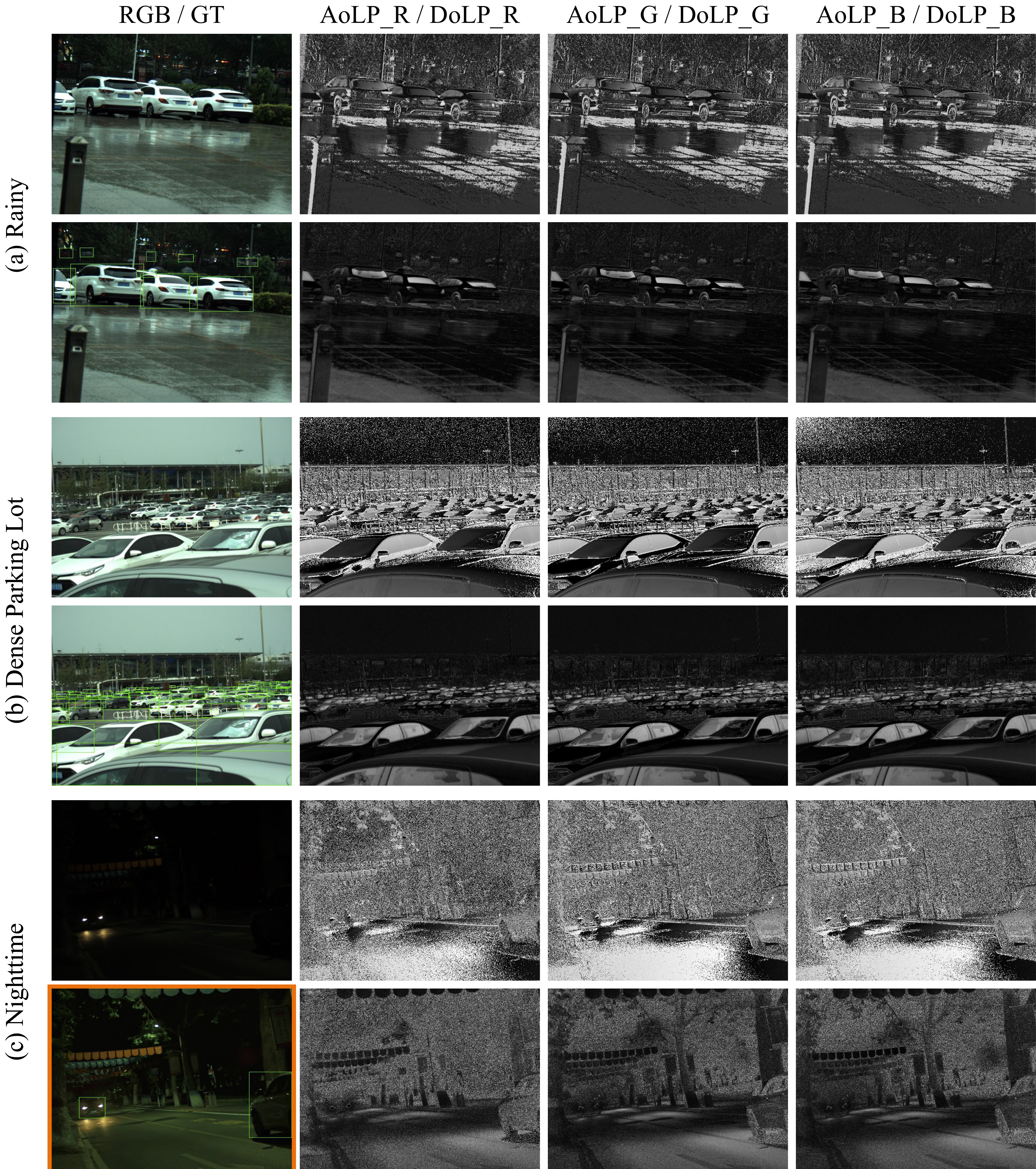}
    \end{center}
    \caption{
    RGBP-Car Examples. The first column displays the RGB intensity (top) and the corresponding annotation (bottom). The next three columns show the AoLP (top) and DoLP (bottom) measurements for the red, green, and blue channels, respectively. From top to bottom are scenes of stopped cars in a rainy parking lot, dense cars in an outdoor parking lot, and driving cars on a clear night road, respectively. 
    (The low-light RGB image is enhanced by ZeroDCE \cite{guo2020zero} (with \orange{orange} frame) for visualization.)}
    \label{fig:samples}
\end{figure}

\begin{figure}[t]
    \centering
    \subfloat[relationship among scenes]{\resizebox{0.22\textwidth}{!}{
        \includegraphics[]{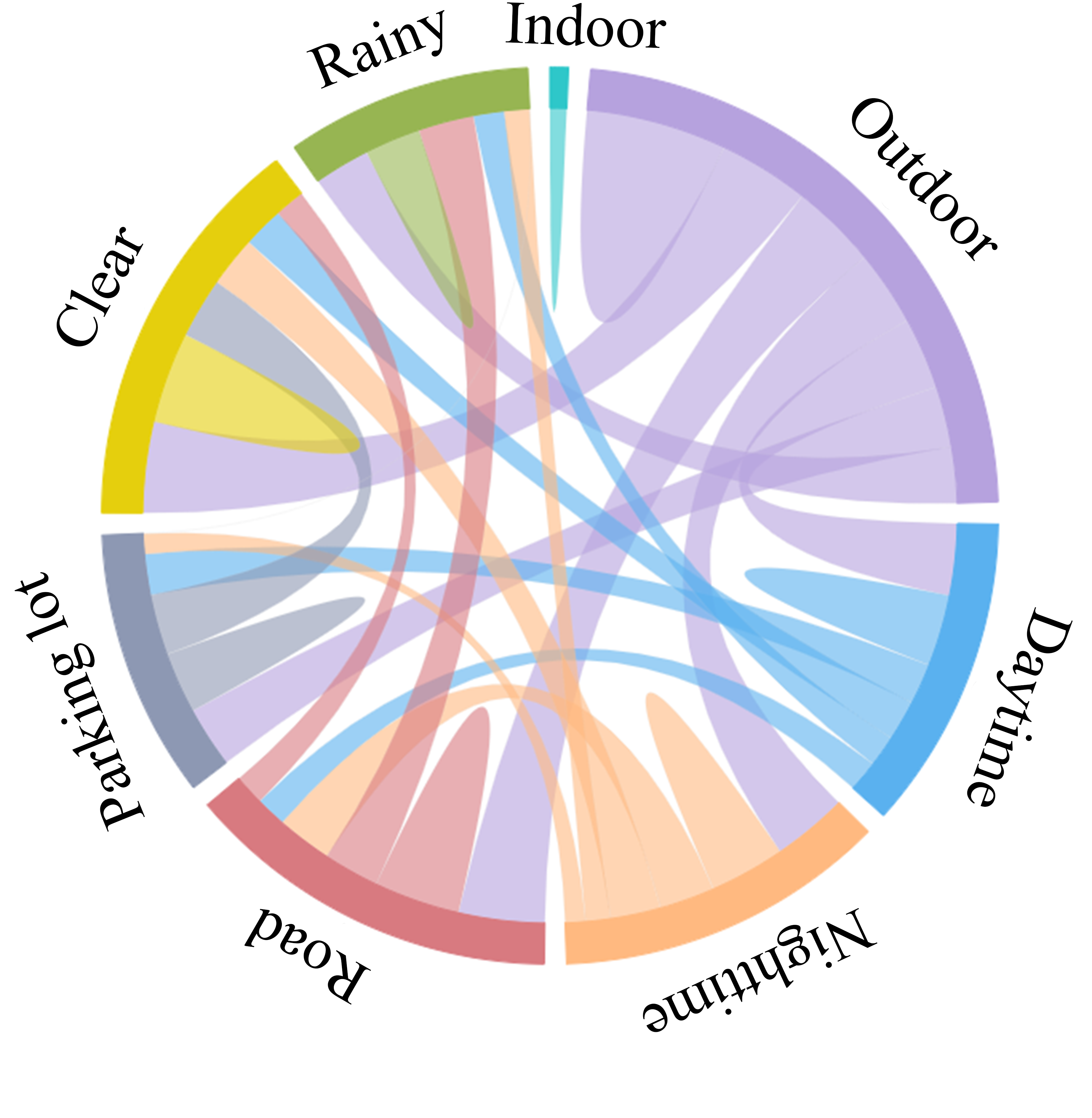}
    }}
    \hspace{4mm}
    \subfloat[car instance $ln$ distribution]{\resizebox{0.22\textwidth}{!}{
        \includegraphics[]{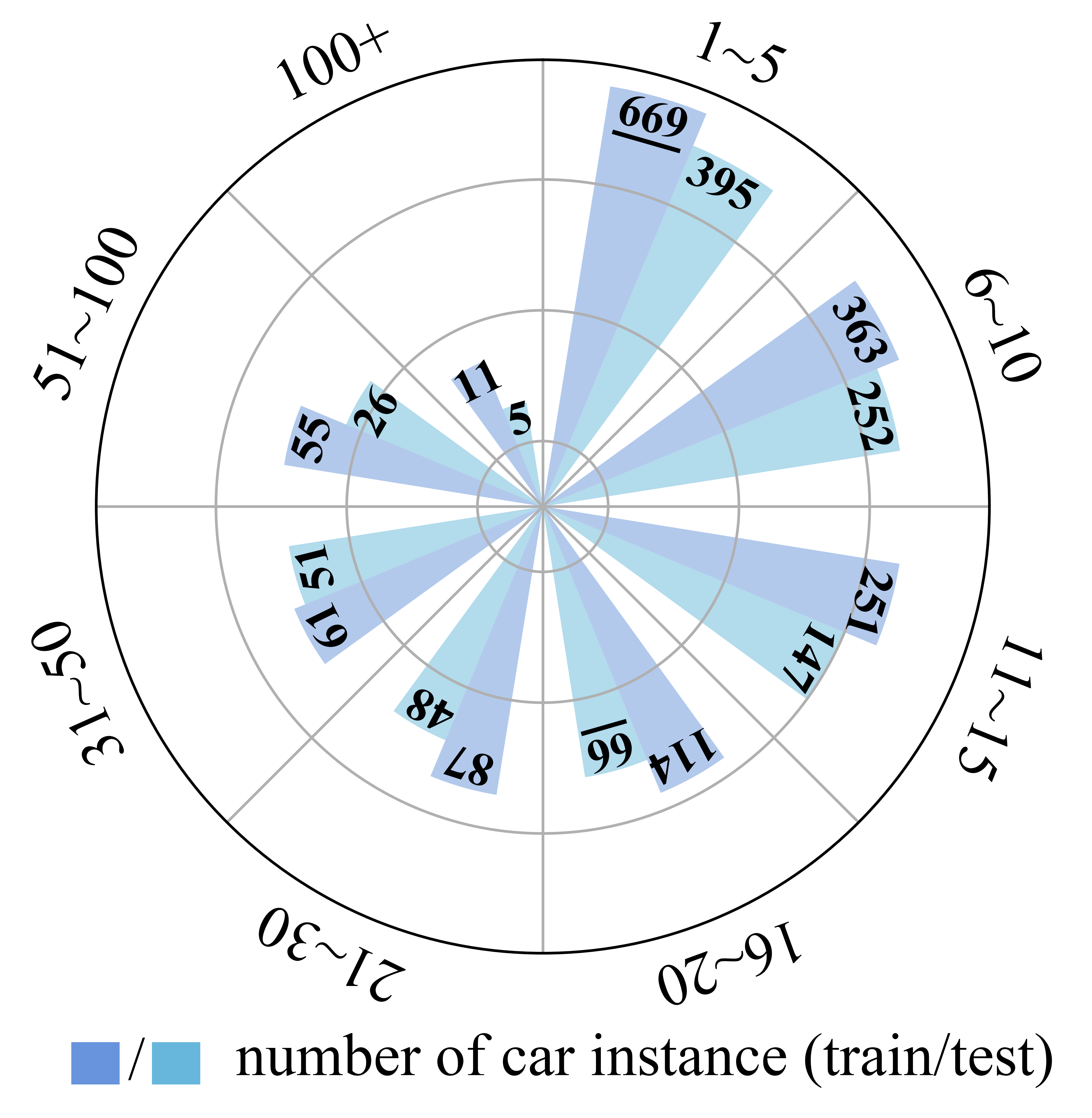}
    }}
    \caption{The images in our RGB-P Car dataset vary in terms of (a) scenarios and (b) the number of car instances.}
    \label{fig:dataset}
\end{figure}

\begin{table}[tp]
\caption{Comparison of existing car detection datasets with polarization measurements.}
\small
\centering
\setlength{\tabcolsep}{2.6pt}
\begin{tabular}{c|c|c|c|c}
\hline\hline
Datasets         & Pol. & \begin{tabular}[c]{@{}c@{}}Pixel\\ align\end{tabular} & \begin{tabular}[c]{@{}c@{}}Num.images \\ Train / Test\end{tabular} & \begin{tabular}[c]{@{}c@{}}Num. cars \\ Train / Test \end{tabular} \\ 
\hline
PolarLITIS       & Mono & $\times$                                                     & \begin{tabular}[c]{@{}c@{}}2569 \\ 1640 / 929 \end{tabular}         & \begin{tabular}[c]{@{}c@{}}17428\\ 6061 / 11367 \end{tabular}    \\
\hline
\textbf{RGBP-Car (Ours)} & Tri  & \checkmark                                                     & \begin{tabular}[c]{@{}c@{}}2601 \\ 1611 / 990 \end{tabular}         & \begin{tabular}[c]{@{}c@{}}31234 \\ 19582 / 11652 \end{tabular}   \\ 
\hline\hline
\end{tabular}
\label{tab:datasetcomp}
\end{table}

\section{RGB-P Car Detection Dataset}
\label{sec:dataset}
We construct the first pixel-aligned RGB-polarization car detection dataset called RGBP-Car with trichromatic polarization measurements. We record cars in diverse traffic scenes using FLIR-Blackfly-S, a polarized color camera that simultaneously obtain pixel-aligned polarization measurements in four linear polarization directions (0$^\circ$, 45$^\circ$, 90$^\circ$, and 135$^\circ$) for each color channel (\textit{i.e.}, R, G, and B). RGBP-Car contains 2601 RGB, AoLP, and DoLP image triplets. Each image has manually labeled bounding boxes indicating the position and size of each car. To ensure the diversity and challenge of our dataset, we take the RGB-P images under different weather conditions (clear and rainy), different lighting conditions (daytime and nighttime), different driving environments (indoor, outdoor, road and parking lot), and different car densities. 
Fig. \ref{fig:samples} gives representative examples and Fig. \ref{fig:dataset} analyzes (a) the relationship among different scenes and (b) the density distribution of car instances. Tab. \ref{tab:datasetcomp} further shows the superiority of our RGBP-Car over existing car detection datasets with polarization measurements.

\section{Assessment}
\label{sec:assessment}
\subsection{Experimental Setup}
We implement our PCDNet in PyTorch \cite{paszke2019pytorch} and train it for 300 epochs with the batch size of 32 on two NVIDIA GeForce RTX 3090 GPUs. We use stochastic gradient descent (SGD) \cite{amari1993backpropagation} with a momentum of 0.937 and a weight decay of $5 \times 10 ^{-4}$ during training. The initial learning rate is set to 0.01 and decayed to 0.001 using a cosine annealing schedule. We initialize PCDNet randomly and load the weights of CSPDarknet53 \cite{wang2020cspnet} pre-trained on ImageNet \cite{imagenet_cvpr09} for the encoder part. To increase the diversity and complexity of the training samples, we apply data augmentations including random cropping, random flipping, and mosaic \cite{redmon2018yolov3}. We use the evaluation metrics of Microsoft COCO \cite{lin2014microsoft} for validation.

\begin{table}[ht]
\caption{Quantitative comparison against state-of-the-art polarization-based detectors ($\star$), single-stage detectors ($\dag$), two-stage detectors ($\ddag$), anchor-based detectors ($\triangle$), anchor-free detectors ($\circ$), and self-supervised method ($\S$).}
\small
\centering
\renewcommand\arraystretch{0.9}
\setlength{\tabcolsep}{2.6pt}
\begin{tabular}{lccccc}
\hline\hline
Methods	&	Pub'Year	&	Backbone	&	AP	&	AP50	&	AP75	\\
\hline
Faster R-CNN$^{\ddag\triangle}$ 	&	NeurIPS'15	&	Res50	&	44.8	&	75.4	&	45.4	\\
SSD$^{\dag\circ}$ 	&	ECCV'16	&	VGG16	&	25.5	&	52.6	&	22.6	\\
Cascade R-CNN$^{\ddag\triangle}$ 	&	CVPR'18	&	Res50	&	45.8	&	73.2	&	47.8	\\
CornerNet$^{\dag\circ}$ 	&	ECCV'18	&	Res50	&	19.8	&	47.4	&	29.6	\\
P-SSD I$^{\star\dag\circ}$ 	&	ITSC'19	&	VGG16	&	25.9 	&	53.1	&	22.7	\\
P-SSD S$^{\star\dag\circ}$ 	&	ITSC'19	&	VGG16	&	23.0 	&	48.9	&	20.1	\\
FCOS$^{\dag\circ}$ 	&	ICCV'19	&	Res50	&	23.1	&	50.9	&	18.4	\\
DH R-CNN$^{\ddag\triangle}$ 	&	CVPR'20	&	Res50	&	32.7	&	65.3	&	28.2	\\
Dynamic R-CNN$^{\ddag\triangle}$ 	&	ECCV'20	&	Res50	&	46.2	&	74.2	&	48.0	\\
EfficientDet$^{\ddag\triangle}$ 	&	CVPR'20	&	D3	&	45.3	&	73.0	&	46.3	\\
VarifocalNet$^{\dag\circ}$  & CVPR'21 & Res50 & 44.2 &	73.5 &	44.4	\\
D-DETR$^{\dag\circ}$ 	&	ICLR'21	&	Res50	&	43.8	&	74.9	&	44.3	\\
DDOD$^{\dag\circ}$ 	&	MM'21	&	Res50	&	43.5	&	73.0	&	43.3	\\
TOOD$^{\dag\triangle}$ 	&	ICCV'21	&	Res50	&	44.3	&	74.3	&	44.6	\\
YOLOX$^{\dag\circ}$ 	&	arXiv'21	&	YOLOX-l	&	54.3	&	82.5	&	56.7	\\
YOLOv7$^{\dag\triangle}$	&	arXiv'22	&	Dark53	&	57.6	&	84.3	&	60.3	\\
RTMDet$^{\dag\circ}$ 	&	arXiv'22	&	RTMDet-l	&	53.9	&	81.4	&	56.7	\\
DINO$^{\dag\circ\S}$ 	&	ICLR'22	&	Res50	&	52.7	&	81.8	&	54.8	\\
YOLOv8$^{\dag\circ}$ 	&	-'23	&	YOLOv8-l	&	56.8	&	83.6	&	59.0	\\
\hline
\textbf{PCDNet$^\star$}	&	\textbf{Ours}	&	Dark53	&	\textbf{58.5}	&	\textbf{85.2}	&	\textbf{61.5}	\\
\hline\hline
\end{tabular}
\label{tab:comparison}
\end{table}

\begin{figure*}[htp]
    \centering
    \begin{center}
        \includegraphics[width=\linewidth,height=10.5cm]{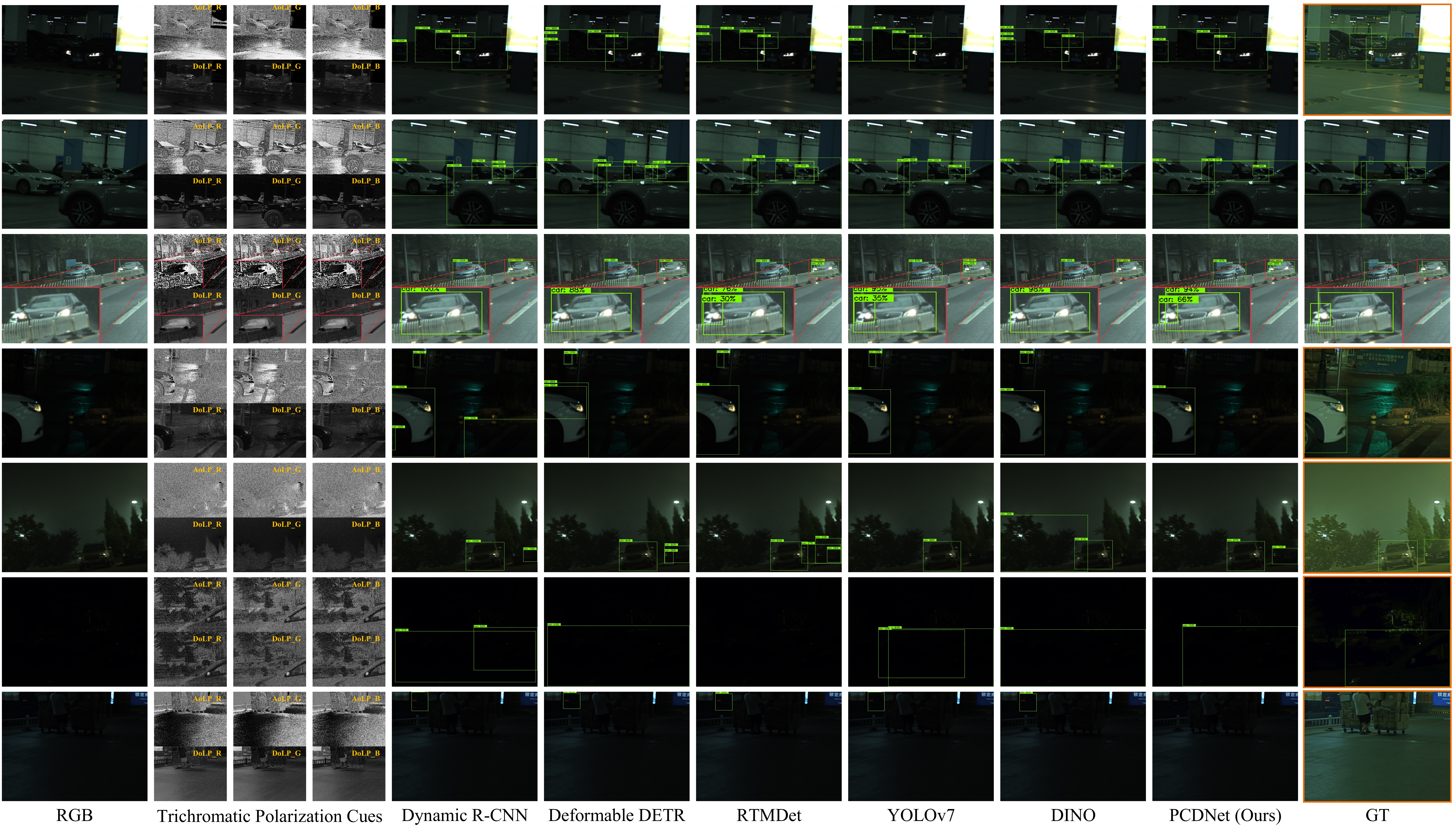}
    \end{center}
    \caption{Qualitative comparison of PCDNet against state-of-the-art detectors retrained on RGB-P Car dataset.} 
    \label{fig:comparison}
\end{figure*}

\subsection{Qualitative and Quantitative Evaluation}
We extensively compare our PCDNet with 19 state-of-the-art methods by retraining and testing all methods on the RGB-P Car dataset using their original settings. The compared methods include two-stage detectors such as EfficientDet \cite{tan2020efficientdet} and the R-CNN family \cite{Ren_2017, Cai_2019, zhang2020dynamic}, and one-stage detectors such as SSD \cite{liu2016ssd}, and YOLO family \cite{ge2021yolox, wang2022yolov7, ultralytics2023yolov8}. These methods also comprise anchor-based methods such as the R-CNN family and YOLOv7 \cite{wang2022yolov7}, and anchor-free methods such as CornerNet \cite{law2018cornernet}, VarifocalNet \cite{zhang2021varifocalnet}, and YOLOv8 \cite{ultralytics2023yolov8}. Some detectors use traditional convolutional networks such as FCOS \cite{tian2019fcos} and RTMDet \cite{lyu2022rtmdet} while others use transformer structures, such as DeformableDETR \cite{zhu2020deformable} and DINO \cite{zhang2022dino} that employs self-supervised learning. We also include the P-SSD \cite{blin2019road} that utilizes polarization information. The quantitative evaluation results are reported in Tab. \ref{tab:comparison}. We can see that our method outperforms all competing state-of-the-art methods. 

Fig. \ref{fig:comparison} further qualitatively demonstrates the benefits of our method: a) in poorly lit indoor parking lots, distinguishing black cars behind pillars is extremely challenging (the first two rows). The compared methods tend to conflate the shadow and the black car (\textit{i.e.}, merging cars on either side of the pillar into a single entity or treating partial views of the car as one object) while our PCDNet can handle such ambiguities; b) in the third example, all methods except our PCDNet fail to detect a partially visible car obstructed by another car or misplace it with the previous car; c) in the fourth example, RGB-based methods wrongly identify distant pedestrians as cars, but our PCDNet method can effectively eliminate such interference with the help of polarization cues; d) the fifth and sixth examples depict black cars in an outdoor parking lot at night which are very hard to be distinguished in the RGB image. Despite the enhancement through ZeroDCE \cite{guo2020zero}, the sixth example remains unclear. By contrast, polarization imaging is robust to low light conditions, enabling our robust car detector PCDNet; and e) the last row shows a virtual car reflected in a mirror located at the upper-left corner of the image. The mirrored virtual car and the rest of the mirror regions exhibit similar and smooth AoLP, providing useful cues for PCDNet to recognize this region as background.

\subsection{Ablation Study}
\begin{figure}[t]
    \centering
    \includegraphics[width=1\linewidth]{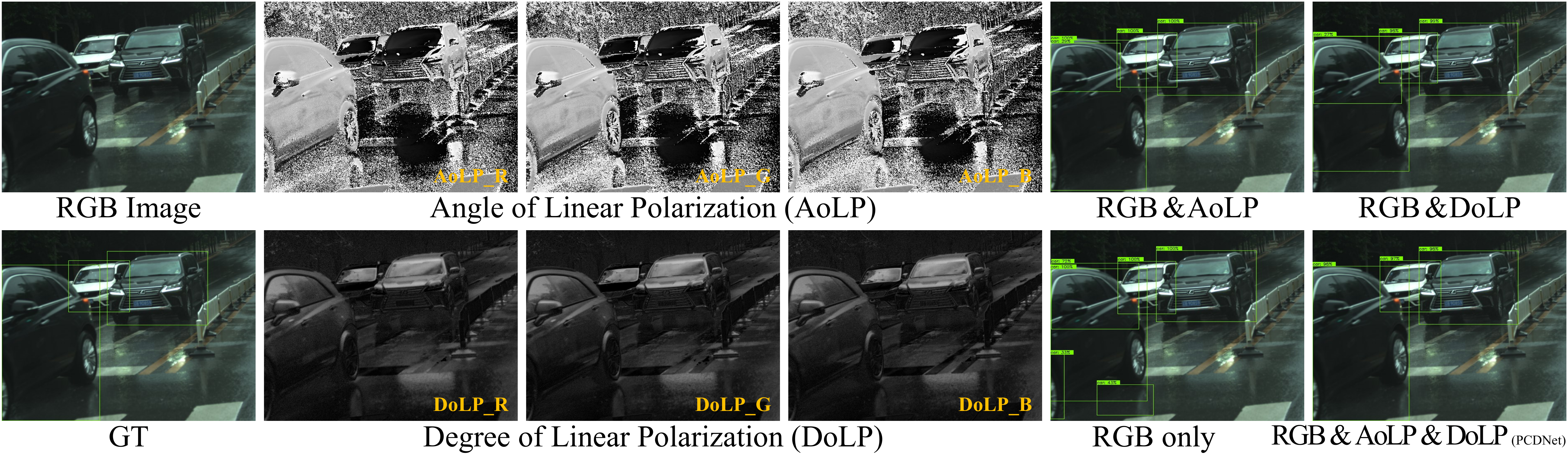}
    \caption{Qualitative comparison of ablation with different inputs. The model with RGB intensity only is susceptible to interference from ghost car caused by water on the road.}
    \label{fig:abl_input}
\end{figure}

\textbf{Impact of Spectral Intensity and Polarization Cues.} We conduct a series of ablation experiments to demonstrate the effects of spectral intensity and polarization cues on car detection (Tab. \ref{tab:abl_input}).
The results show that: a) combining different forms of polarization cues with RGB as the input of PCDNet can improve the car detection accuracy (\textit{C}, \textit{D}, \textit{F}, \textit{G}, \textit{K} and \textit{L} are higher than \textit{B}); b) DoLP cues have a greater impact than AoLP cues (\textit{D}, \textit{J} and \textit{L} are better than \textit{C}, \textit{I} and \textit{K}, respectively); c) stacking AoLP and DoLP on RGB in the channel dimension does not boost performance (\textit{E} is slightly lower than \textit{B}), possibly because the characteristic gap between different modalities hinders effective features extraction; d) spectral intensity and polarization are more beneficial than monochromatic intensity and polarization for car detection (comparing paired \textit{B} and \textit{H}, \textit{C} and \textit{K}, \textit{D} and \textit{L}, \textit{I} and \textit{K}, \textit{J} and \textit{L}); e) enhancing RGB image via ZeroDCE \cite{guo2020zero} is less effective than introducing polarization (\textit{M} performs worse than \textit{C}-\textit{G}, \textit{K} and \textit{L}).
Fig. \ref{fig:abl_input} provides visual support for these observations.

\begin{table}[t]
\small
\centering
\caption{Quantitative comparisons of ablation with different inputs. ``stacked I'' denotes the stacked intensity measurements with a linear polarization angle of 0$^{\circ}$, 45$^{\circ}$ and 135$^{\circ}$ and ``stacked S'' refers to the stacked Stokes elements S0, S1 and S2 \cite{blin2019road}.}
\begin{tabular}{clccc}
\hline\hline
	&	PCDNet Input	&	AP	&	AP50	&	AP75	\\
 \hline
\textit{A}	&	RGB, AoLP and DoLP (original)	&	58.5 	&	85.2 	&	61.5 	\\
\hline
\textit{B}	&	RGB only	&	57.6 	&	84.3 	&	60.2 	\\
\textit{C}	&	RGB and AoLP	&	58.0 	&	84.6 	&	60.7 	\\
\textit{D}	&	RGB and DoLP	&	58.3 	&	85.4 	&	61.1 	\\
\textit{E}	&	stacked RGB, AoLP and DoLP	&	57.5 	&	84.3 	&	59.9 	\\
\textit{F}	&	RGB and stacked I	&	58.0 	&	84.1 	&	61.0 	\\
\textit{G}	&	RGB and stacked S	&	57.8 	&	84.8 	&	60.4 	\\
\textit{H}	&	Gray only	&	57.4 	&	84.3 	&	60.0 	\\
\textit{I}  &   Gray and mono AoLP & 57.5 & 84.5 & 60.5 \\
\textit{J}  &   Gray and mono DoLP & 57.6 & 84.9 & 60.1 \\
\textit{K}	&	RGB and mono AoLP	&	57.9 	&	84.6 	&	60.5 	\\
\textit{L}	&	RGB and mono DoLP	&	58.2 	&	84.9 	&	60.6 	\\
\textit{M}  &   Enhanced RGB & 57.4 & 84.0 & 60.0 \\
\hline\hline
\end{tabular}
\label{tab:abl_input}
\end{table}

\textbf{Influence of PCDNet Components.}
First, we investigate the performance of different strategies for fusing AoLP and DoLP inputs. From Tab. \ref{tab:abl_module}(\textit{A}-\textit{D}), we observe that our PI module is more effective than the simple fusion methods including concatenation, addition and element-wise multiplication.
Second, by removing MP module \ref{tab:abl_module}(\textit{E}) from the original PCDNet (A), the detection performance declines. This demonstrates that exploring the polarized material features of cars across all learning samples is useful. We also explore the influence of applying MSP and MCP on different levels of features. The results in Tab. \ref{tab:abl_module}(\textit{A},\textit{F}-\textit{G}) show that applying MSP on shallower features and MCP on deeper features can yield better performance.
Finally, we validate the effectiveness of CDDQ module.
Removing the CDDQ module (\textit{I}) from PCDNet (\textit{A}), which causes the feature extraction processes of the RGB and polarization to be independent from each other, leads to the performance drop. We also demonstrate the benefits of the CWDA and SDMD in the CDDQ module by removing either of them (\textit{J} and \textit{K}). 

\begin{table}[t]
\small
\centering
\caption{Quantitative comparisons of ablation with different modules demonstrate that all component of PCDNet contributes to the overall performance. We used sequences of three letters separated by '-' and enclosed in parentheses to represent different combinations of MSP and MCP.}
\begin{tabular}{clccc}
\hline\hline
	&	Ablation	&	AP	&	AP50	&	AP75	\\
 \hline
\textit{A}	&	PCDNet (original)	&	58.5 	&	85.2 	&	61.5 	\\
\hline
\textit{B}	&	Input RGB and [AoLP DoLP]	&	58.2 	&	85.4 	&	60.9 	\\
\textit{C}	&	Input RGB and AoLP+DoLP	&	58.1 	&	84.8 	&	60.5 	\\
\textit{D}	&	Input RGB and AoLP*DoLP	&	58.1 	&	84.8 	&	60.5 	\\
\hline
\textit{E}	&	A \textit{w/o} MP	&	56.9 	&	84.2 	&	59.2 	\\
\textit{F}	&	A \textit{w/} M(S-S-S)P	&	58.2 	&	85.2 	&	60.8 	\\
\textit{G}	&	A \textit{w/} M(S-C-C)P	&	58.2 	&	85.0 	&	60.9 	\\
\textit{H}	&	A \textit{w/} M(C-C-C)P	&	58.1 	&	85.0 	&	61.1 	\\
\hline
\textit{I}	&	A \textit{w/o} CDDQ	&	58.0 	&	84.7 	&	60.8 	\\
\textit{J}	&	A \textit{w/o} SDMD	&	58.2 	&	85.2 	&	60.8 	\\
\textit{K}	&	A \textit{w/o} CWDA	&	58.3 	&	85.1 	&	61.1 	\\
\hline\hline
\end{tabular}
\label{tab:abl_module}
\end{table}

\subsection{Limitations}

\begin{figure}[t]
    \centering
    \includegraphics[width=1\linewidth]{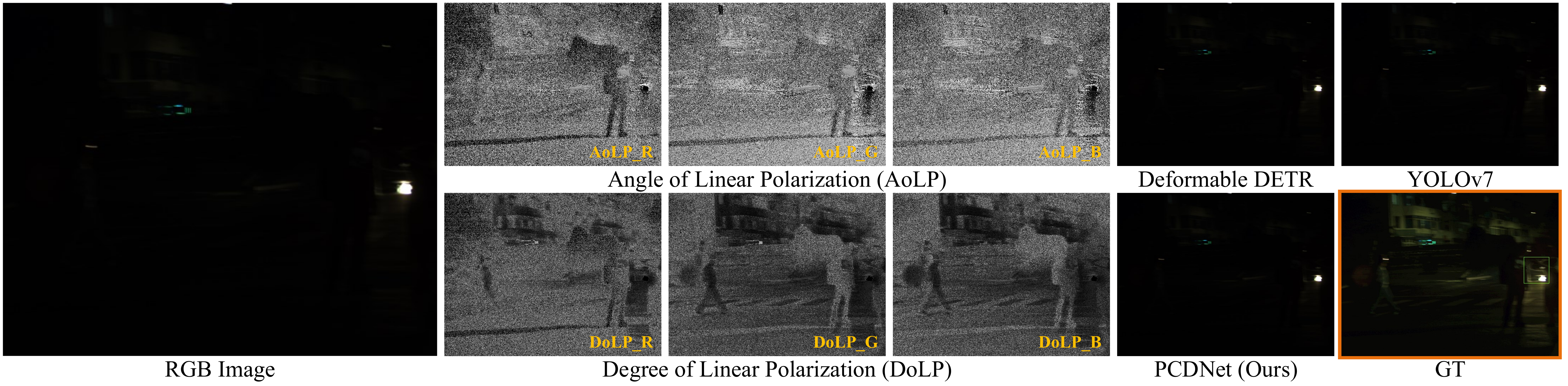}
    \caption{PCDNet has limited ability to handle extreme HDR or heavy motion blur cases.}
    \label{fig:failure}
\end{figure}

When both the RGB intensity and the polarization measurement yield weak car signals, our method's effectiveness declines. Specifically, in low-light scenarios, when a car approaches on an unlit road, the strong light from its headlights can create a ``hole'' in both the RGB and polarization and obscure the entire car. We illustrate such an example in Fig. \ref{fig:failure} where the extreme HDR and heavy motion blur in the captured image limit its depiction of both RGB and polarization. In these challenging scenarios, prior RGB-based methods and even human vision are powerless.

\section{Conclusion}
\label{sec:conclusion}
In this paper, we present PCDNet, the first solution that leverages both RGB intensities and trichromatic angle/degree of linear polarization (AoLP/DoLP) cues for robust car detection in challenging scenarios. PCDNet comprises three key modules: the Polarization Integration (PI) module, the Material Perception (MP) module, and the Cross-Domain Demand Query (CDDQ) module. The PI module fuses AoLP and DoLP to generate a comprehensive and semantically meaningful polarization representation. The MP module explores the polarization/material properties of cars across different learning samples and the CDDQ module proactively integrates RGB features and polarization representations in a request-and-complement manner. Extensive experiments show that PCDNet outperforms existing methods, especially in challenging scenarios. We also introduced a new pixel-aligned RGB-P car detection dataset covering diverse scenarios, which can promote the use of polarization in relevant visual tasks.

\section{Acknowledgements}
This work was supported in part by National Key Research and Development Program of China (2022ZD0210500/2021ZD0112400), the National Natural Science Foundation of China under Grants 62102058/62272081/61972067/62332019, the Distinguished Young Scholars Funding of Dalian (No. 2022RJ01), and the open funding of State Key Laboratory of Structural Analysis for Industrial Equipment.

\bibliography{aaai24}

\end{document}